\newcommand{\CLASSINPUTtoptextmargin}{60pt}
\newcommand{\CLASSINPUTbottomtextmargin}{120pt}

\documentclass[10pt,a4paper]{IEEEtran}

\usepackage{graphicx}
\usepackage{amsmath}
\usepackage{amssymb}
\usepackage{multirow}
\usepackage[pagebackref=true,breaklinks=true,letterpaper=true,colorlinks,bookmarks=false]{hyperref}
\usepackage{authblk}

\begin{document}

	\title{Fast Localization of Facial Landmark Points}

	\author[$\dagger$]{Nenad Marku\v{s}}
	\author[$\dagger$]{Miroslav Frljak}
	\author[$\dagger$]{Igor S. Pand\v{z}i\'{c}}
	\author[$\ddagger$]{J\"orgen Ahlberg}
	\author[$\ddagger$]{Robert Forchheimer}

	\affil[$\dagger$]
	{
		University of Zagreb,
		Faculty of Electrical Engineering and Computing,
		Unska 3, 10000 Zagreb, Croatia
	}
	\affil[$\ddagger$]
	{
		Link\"{o}ping University,
		Department of Electrical Engineering,
		SE-581 83 Link\"{o}ping, Sweden
	}

	\maketitle

	\begin{abstract}
		Localization of salient facial landmark points, such as eye corners or the tip of the nose, is still considered a challenging computer vision problem despite recent efforts. 
		This is especially evident in unconstrained environments, i.e., in the presence of background clutter and large head pose variations.
		Most methods that achieve state-of-the-art accuracy are slow, and, thus, have limited applications.
		We describe a method that can accurately estimate the positions of relevant facial landmarks in real-time even on hardware with limited processing power, such as mobile devices.
		This is achieved with a sequence of estimators based on ensembles of regression trees.
		The trees use simple pixel intensity comparisons in their internal nodes and this makes them able to process image regions very fast.
		We test the developed system on several publicly available datasets and analyse its processing speed on various devices.
		Experimental results show that our method has practical value.
	\end{abstract}

	\section{Introduction}
			The human face plays an essential role in everyday interaction, communication and other routine activities.
			Thus, automatic face analysis systems based on computer vision techniques open a wide range of applications.
			Some include biometrics, driver assistance, smart human-machine interfaces, virtual and augmented reality systems, etc.
			This serves as a strong motivation for developing fast and accurate automatic face analysis systems.

			In this paper we describe a novel method for estimating the positions of salient facial landmark points from an image region containing a face.
			This is achieved by extending our previous work in eye pupil localization and tracking \cite{puploc}.
			The developed prototype achieves competitive accuracy and runs in real-time on hardware with limited processing power, such as mobile devices.
			Additionally, one of the main advantages of our approach is its simplicity and elegance.
			For example, we completely avoid image preprocessing or the computation of special structures for fast feature extraction, such as integral images and HOG pyramids:
			the method works on raw pixel intensities.

		\subsection{Relevant recent work}
			Significant progress has been achieved recently in the area of facial landmark localization.
			The methods considered to be state-of-the-art are described in \cite{sota-belhumeur, sota-explicitshape, sota-neural}.
			The approach described by Belhumeur et al. \cite{sota-belhumeur} outperformed previously reported work by a large margin.
			It combines the output of SIFT-based face part detectors with a non-parametric global shape model for the part locations.
			The main drawback with this approach is its low processing speed.
			Cao et al. \cite{sota-explicitshape} described a regression-based method for face alignment.
			Their idea is to learn a function that directly maps the whole facial shape from the image as a single high-dimensional vector.
			The inherent shape constraint is naturally encoded in the output.
			This makes it possible to avoid parametric shape models commonly used by previous methods.
			As this is a tough regression problem, they fit the shape in a coarse-to-fine manner using a sequence of fern\footnote{A simplified decision tree, see \cite{ferns}.} ensembles with shape-indexed pixel intensity comparison features.
			The developed system is both fast and accurate.
			The system developed by Sun et al. \cite{sota-neural} is based on a deep convolutional neural network trained to estimate the positions of five facial landmarks.
			Additionally, simpler networks are used to further refine the results.
			The authors report state-of-the-art accuracy results.
			Recently, deep neural networks started to outperform other methods on many machine learning benchmarks (for example, see \cite{hinton}).
			Thus, this is not at all surprising.
			However, neural networks are usually slow at runtime as they require a lot of floating point computations to produce their output, which is particularly problematic on mobile devices.
			Chevallier et al. \cite{slicnonama} described a method similar to the one we present in this paper.
			We address this later in the text.

	\section{Method}
		The basic idea is to use a multi-scale sequence of regression tree-based estimators to infer the position of each facial landmark point within a given face region.
		We assume that this region is known in advance.
		This does not pose a problem since very efficient and accurate face detectors exist
		(including our own work \cite{pico}).
		In contrast to most prior work, we treat each landmark point separately, disregarding the correlation between their positions.
		Of course, a shape constraint could be enforced in the post-processing step and there are many methods to achieve this.
		We have decided to exclude this step in order to focus on landmark localization itself.
		We explain the details of the method in the rest of this section and compare it with previous approaches.

		\subsection{Regression trees based on pixel intensity comparisons}
			To address the problem of image based regression, we use an optimized binary decision tree with pixel intensity comparisons as binary tests in its internal nodes.
			Variations of this approach have already been used by other researchers to solve certain computer vision problems (for example, see \cite{kinect, tld, ferns}).
			We define a pixel intensity comparison binary test on image $I$ as
			$$
				\text{bintest}(I; \mathbf{l}_1, \mathbf{l}_2) =
				\begin{cases}
					0,	&	I(\mathbf{l}_1)\leq I(\mathbf{l}_2)	\\
					1,	&	\text{otherwise}
					,
				\end{cases}
			$$
			where $I(\mathbf{l}_i)$ is the pixel intensity at location $\mathbf{l}_i$.
			Locations $\mathbf{l}_1$ and $\mathbf{l}_2$ are in normalized coordinates, i.e., both are from the set $[-1, +1]\times[-1, +1]$.
			This means that the binary tests can easily be resized if needed.
			Each terminal node of the tree contains a vector that models the output.
			In our case, this vector is two-dimensional since we are interested in estimating the landmark position within a given image region.

			The construction of the tree is supervised.
			The training set consists of images annotated with values in $\mathbb{R}^2$.
			In our case, these values represent the location of the landmark point in normalized coordinates.
			The parameters of each binary test in internal nodes of the tree are optimized in a way to maximize clustering quality obtained when the incoming training data is split by the test.
			This is performed by minimizing
			\begin{equation}
				Q =
				\sum_{\mathbf{x}\in C_0} \|\mathbf{x} - \bar{\mathbf{x}}_0\|_2^2 + \sum_{\mathbf{x}\in C_1} \|\mathbf{x} - \bar{\mathbf{x}}_1\|_2^2
				\label{eq:mse}
				,
			\end{equation}
			where $C_0$ and $C_1$ are clusters that contain landmark point coordinates $\mathbf{x}\in\mathbb{R}^2$ of all face regions for which the outputs of binary test were $0$ and $1$, respectively.
			The vector $\bar{\mathbf{x}}_0$ is the mean of $C_0$ and $\bar{\mathbf{x}}_1$ is the mean of $C_1$.
			As the set of all pixel intensity comparisons is prohibitively large, we generate only a small subset\footnote{$128$ in our implementation.} during optimization of each internal node by repeated sampling of two locations from a uniform distribution on a square $[-1, +1]\times[-1, +1]$.
			The test that achieves the smallest error according to equation \ref{eq:mse} is selected.
			The training data is recursively clustered in this fashion until some termination condition is met.
			In our setup, we limit the depth of our trees to reduce training time, runtime processing speed and memory requirements.
			The output value associated with each terminal node is obtained as the weighted average of ground truths that arrived there during training.

			It is well known that a single tree will most likely overfit the training data.
			On the other hand, an ensemble of trees can achieve impressive results.
			A popular way of combining multiple trees is the gradient boosting procedure \cite{gradboost}.
			The basic idea is to grow trees sequentially.
			Each new one added to the ensemble is learned to reduce the remaining training error further.
			Its output is shrunk by a scalar factor called shrinkage, a real number in the set $(0, 1]$, that plays a role similar to the learning rate in neural networks training.
			We set this value to $0.5$ in our experiments.

		\subsection{Estimating the position of a landmark point}
			We have observed that accuracy and robustness of the method critically depend on the scale of the rectangle within which we perform the estimation.
			If the rectangle is too small, we risk that it will not contain the facial landmark at all due to the uncertainty introduced by face tracker/detector used to localize the rectangle.
			If the rectangle is big, the detection is more robust but accuracy suffers.
			To minimize these effects, we learn multiple tree ensembles, each for estimation at different scale.
			The method proceeds in a recursive manner, starting with an ensemble learned for largest scale.
			The obtained intermediate result is used to position the rectangle for the next ensemble in the chain.
			The process continues until the last one is reached.
			Its output is accepted as the final result.
			This was inspired by the work done by Ong et al. \cite{multiscalelinpred}.

			The output of regression trees is noisy and can be unreliable in some frames, especially if the video stream is supplied from a low quality camera.
			This can be attributed to variance of the regressor as well as to the simplicity of binary test at internal nodes of the trees: Pixel footprint size changes significantly with variations in scale of the eyes and we can expect problems with aliasing and random noise.
			These problems can be reduced during runtime with random perturbations \cite{cascadedposeregression}.
			The idea is to sample multiple rectangular regions at different positions and scales around the face and estimate the landmark point position in each of them.
			We obtain the result as the median over the estimations for each spatial dimension.

			We would like to note that Chevallier et al. \cite{slicnonama} described a similar method for face alignment.
			The main difference is that they use Haar-like features instead of pixel intensity comparisons to form binary tests in internal nodes of the trees.
			Also, they do not perform random perturbations in runtime.
			This is presumably not needed with Haar-like features as they are based on region averaging which is equivalent to low pass filtering and this makes them more robust to aliasing and noise.

	\section{Experimental analysis}
		We are interested in evaluating the usefulness of the method in relevant applications.
		Thus, we provide experimental analysis of its implementation in the C programming language.
		We compare its accuracy with the reported state-of-the-art and modern commercial software.
		Also, we analyse its processing speed and memory requirements.

		\subsection{Learning the estimation structures}
			We use the LFW dataset \cite{fanelli} and the one provided by Visage Technologies (\url{http://www.visagetechnologies.com/}).
			Both consist of face images with annotated coordinates of facial landmarks.
			These include the locations of eyebrows, nose, upper and lower lip, mouth and eye corners.
			Overall, the total number of annotated faces is around $15\;000$.
			We intentionally introduce position and scale perturbations in the training data in order to make our system more robust.
			We extract a number of samples from each image by randomly perturbing the bounding box of the face.
			Furthermore, as faces are symmetric, we double the size of the training data by mirroring the images and modifying the landmark point coordinates in an appropriate manner.
			This process results in a training set that consists of approximately $10\;000\;000$ samples.

			Each landmark point position estimation structure is learned independently in our framework.
			We have empirically found that $6$ stages with $20$ trees of depth equal to $9$ give good results in practice.
			The ensemble of the first stage is learned to estimate the position of a particular landmark point from the bounding box of the face.
			Each next stage is learned on a training set generated by shrinking the bounding box by $0.7$ and repositioning its center at the position output by the ensemble of the previous stage.
			This process proceeds until the last stage is reached.
			The learning of the whole estimation structure for a single landmark point takes around one day on a modern desktop computer with $4$ cores and $16$ GB of memory.

		\subsection{Accuracy analysis on still images}
			We use the BioID \cite{bioid} and LFPW \cite{sota-belhumeur} face datasets to evaluate the accuracy in still images.
			The normalized error \cite{bioid} is adopted as the accuracy measure for the estimated landmark point locations:
			\begin{equation}\label{eq:nerr}
				e=
				\frac{1}{N}\sum_{n=1}^N \frac{D_n}{D}
				,
			\end{equation}
			where $N$ is the number of facial landmarks, $D$ is the distance between the eyes and $D_n$ is the Euclidean distance between the estimated landmark position and the ground truth.
			The accuracy is defined as the fraction of the estimates having an error smaller than a given number.
			Roughly, an error of $0.25$ corresponds to the distance between the eye center and the eye corners, $0.1$ corresponds to the diameter of the iris, and $0.05$ corresponds to the diameter of the pupil.

\begin{figure}
				\center
				\includegraphics[scale=0.5]{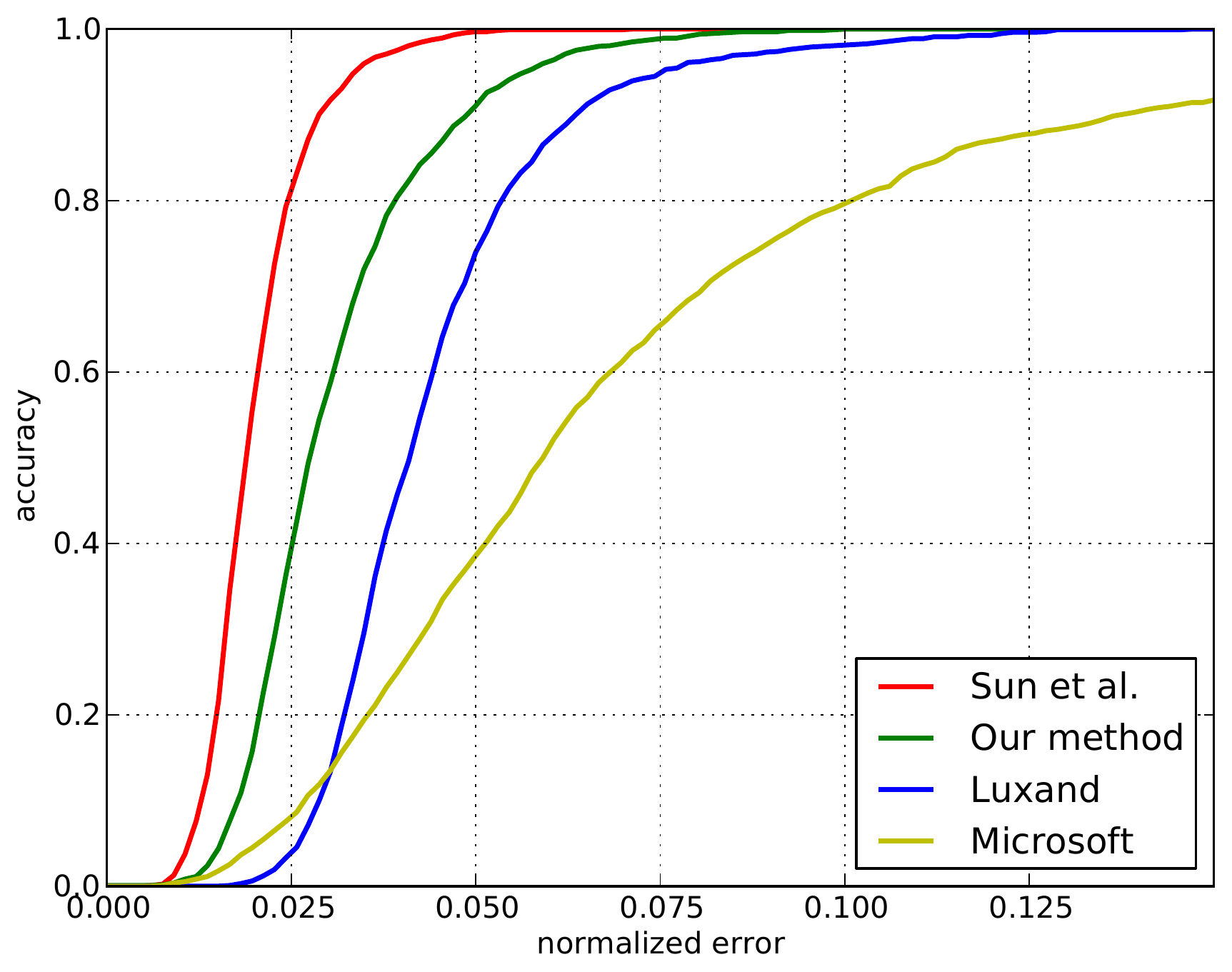}

				\caption
				{
					Accuracy curves on the BioID dataset.
				}
				\label{fig:rec-bioid}
			\end{figure}

\begin{figure}
				\center
				\includegraphics[scale=0.5]{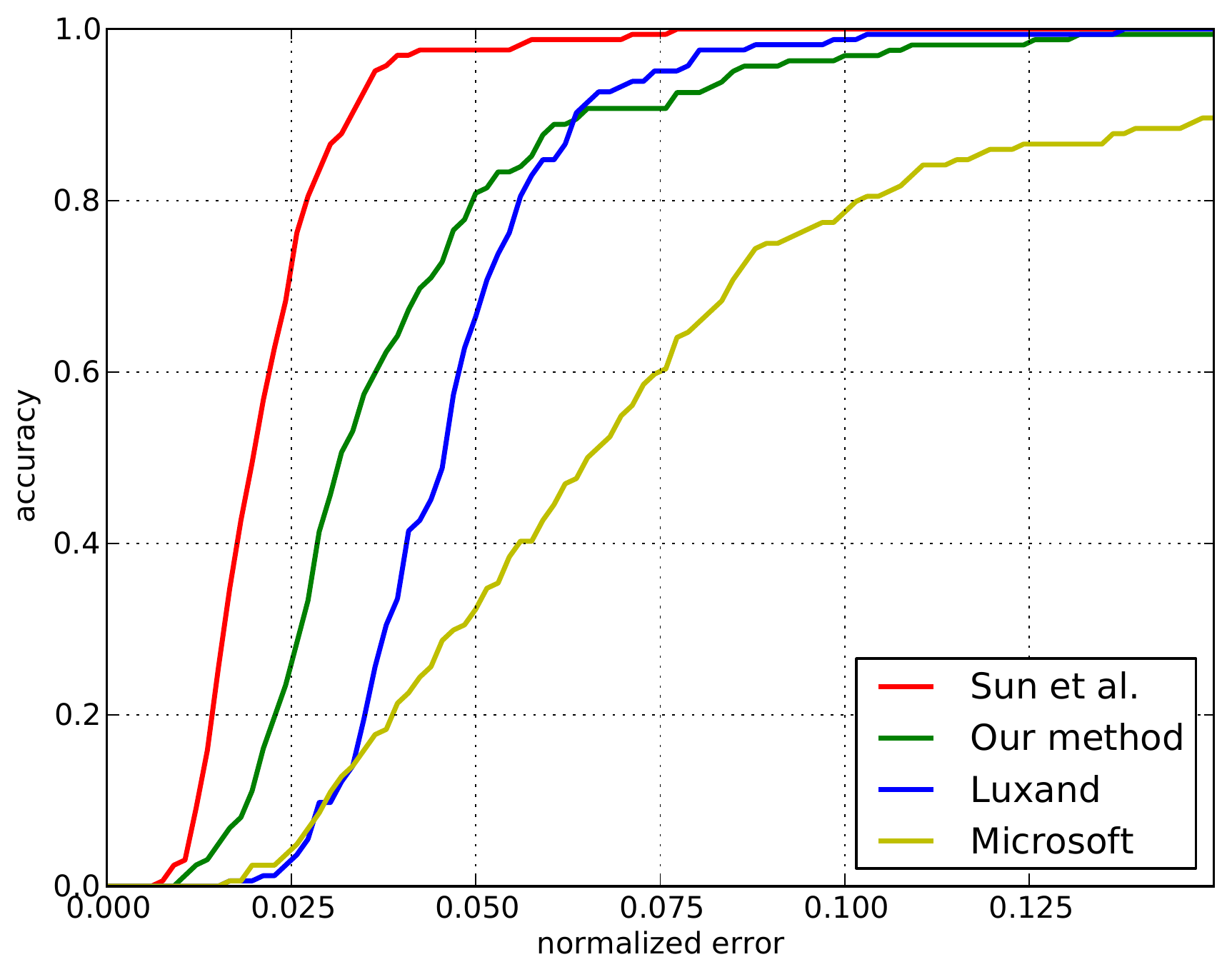}

				\caption
				{
					Accuracy curves on the LFPW dataset.
				}
				\label{fig:rec-lfpw}
			\end{figure}

			First, we use the BioID and LFPW datasets to compare our system to the state-of-the-art method based on convolutional neural networks \cite{sota-neural} and two modern commercial systems, one provided by Microsoft \cite{winphone} and the other by Luxand \cite{luxand}.
			We follow the protocol from \cite{sota-neural} to obtain normalized errors for five facial landmarks (eye centers, mouth corners and tip of the nose).
			The results are reported in figures \ref{fig:rec-bioid} and \ref{fig:rec-lfpw}.
			We can see that our method outperforms both commercial systems in accurate landmark point localization ($e\approx 0.05$) but the neural network-based system is clearly the best.
			We could not include the method by Cao et al. \cite{sota-explicitshape} in this comparison since no code/binaries are available to reproduce the results.
			We also excluded the results published by Chevallier et al. \cite{slicnonama} since they used the evaluation methodology where they partitioned BioID in two parts: one for training and the other for accuracy analysis.
			It has been argued that this evaluation methodology is flawed since the learning procedure overfits some particular features present only in the used dataset and thus yields performance that is not representative in the general case \cite{datasetbias}.

\begin{figure}
				\center
				\includegraphics[scale=0.5]{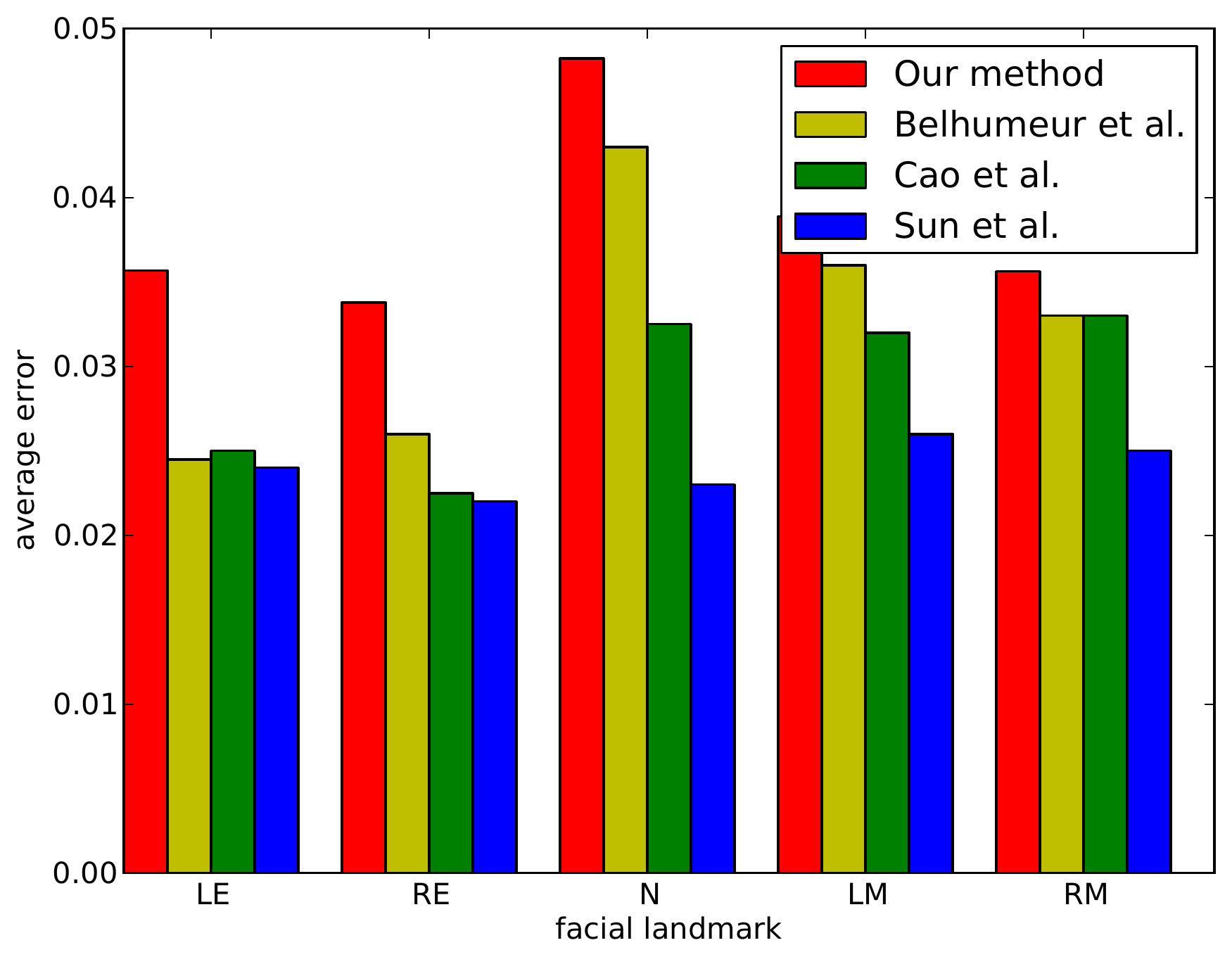}
				\caption
				{
					Average accuracy comparison on the LFPW test set (249 images).
				}
				\label{fig:avgerrs-lfpw}
			\end{figure}

			In order to compare our method also with the methods excluded from the first experiment, we performed a second comparison.
			This one is based on average errors reported on the LFPW dataset
			(accuracy curves could not be obtained due to the lack of data in \cite{sota-belhumeur} and \cite{sota-explicitshape}).
			The average error for $5$ facial landmarks can be seen in Figure \ref{fig:avgerrs-lfpw}.
			As the average error is sensitive to outliers and LFPW faces vary greatly in pose and occlusions, our method performs worse than other approaches that use some form of shape constraint.
			Nevertheless, we can see that, on average, the landmark positions estimated by our system are within the pupil diameter from the ground truth.

		\subsection{Tracking facial features}
\begin{figure}
				\center
				\includegraphics[scale=0.5]{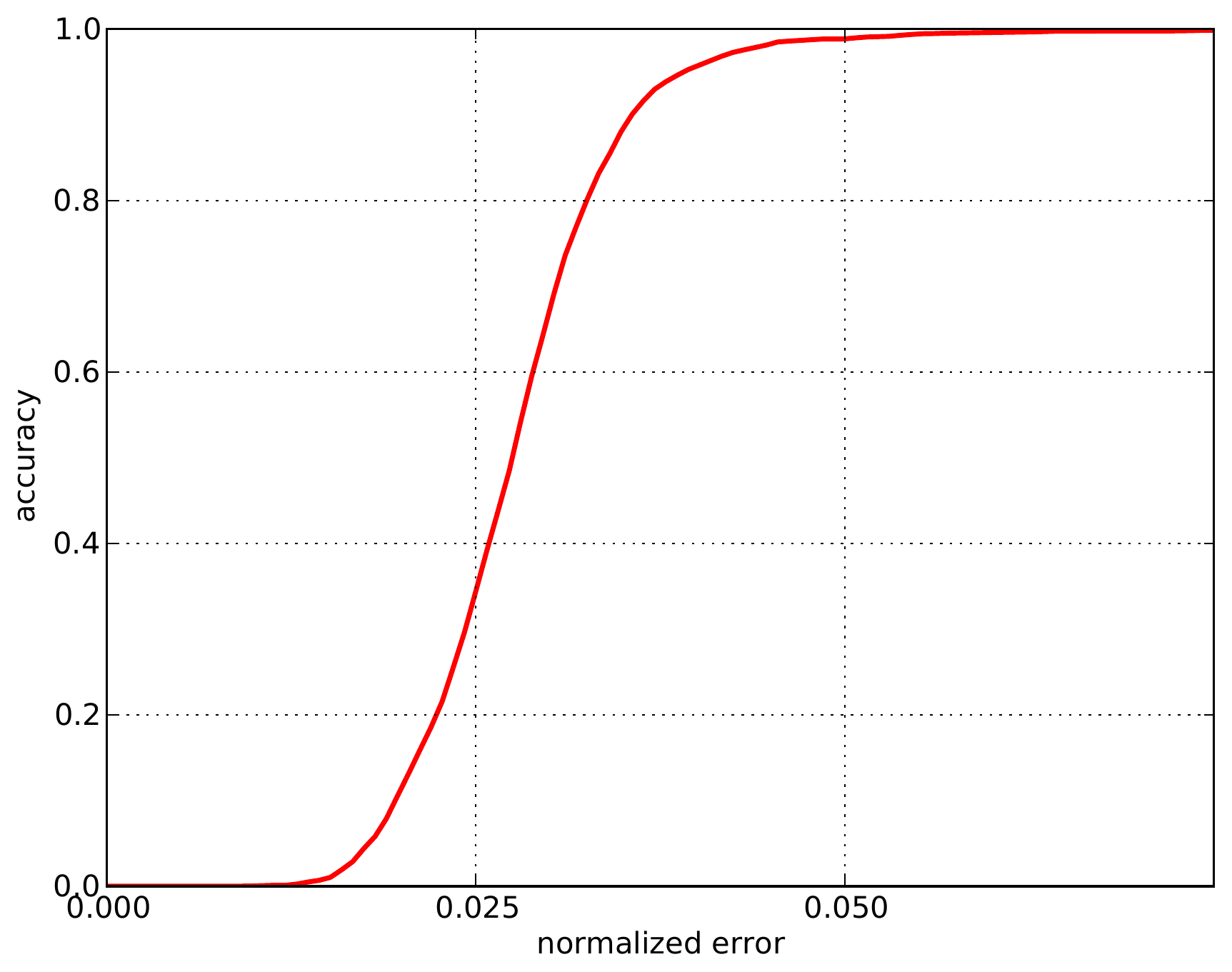}

				\caption
				{
					Accuracy curve for The Talking Face Video.
				}
				\label{fig:rec-frank}
			\end{figure}

			We use the Talking Face Video \cite{frank} to evaluate our system quantitatively in real-time applications.
			The video contains 5000 frames taken from a video of a person engaged in conversation.
			A number of facial landmarks were annotated semi-automatically for each frame with an active appearance model trained specifically for the person in the video.
			These annotations include the locations of eye centers, mouth corners and the tip of the nose.
			The normalized error averaged over the video sequence obtained by our system was equal to $0.028$.
			Accuracy curve can be seen in Figure \ref{fig:rec-frank}.
			These results show that most of the time our system estimated the positions of facial landmarks with high accuracy.

		\subsection{Processing speed and memory requirements}
			Processing speeds obtained by our system on various devices can be seen in Table \ref{tbl:speed}.
			Our system uses a single CPU core although the computations can easily be parallelized.
			\begin{table}
				\center
				\begin{tabular}{| c | c | c |}
					\hline
					Device	&	CPU	&	Time [ms]	\\
					\hline
					\hline
					PC1	&	3.4GHz Core i7-2600	&	$1.3$	\\
					\hline
					PC2	&	2.53GHz Core 2 Duo P8700	&	$2.1$	\\
					\hline
					iPhone 5	&	1.3GHz Apple A6	&	$5.0$	\\
					\hline
					iPad 2	&	1GHz ARM Cortex-A9	&	$8.8$	\\
					\hline
					iPhone 4S	&	800MHz ARM Cortex-A9	&	$10.9$	\\
					\hline
				\end{tabular}
				\caption
				{
					Average times required to align $5$ facial landmarks.
				}
				\label{tbl:speed}
			\end{table}
			Both Cao et al. \cite{sota-explicitshape} and Sun et al. \cite{sota-neural} vaguely\footnote{For example, we are not sure if they used multi-core processing in runtime (both papers mention it at some point).} report processing speeds on modern CPUs: their systems localize $29$ and $5$ facial landmarks, respectively, in $5$ and $120$ [ms], respectively.

			Each landmark position estimator consisting of $120$ trees, each of depth $9$, requires around $700$ kB of memory.
			In our opinion, these relatively large memory requirements are one of the drawbacks with our approach as they are inconvenient for some applications, such as face tracking in web browsers or on mobile devices.
			The problem can be addressed by quantizing the outputs in the leafs of each tree.
			In the current implementation, we represent each output with two $32$-bit floating point values.

		\subsection{Qualitative results}
\begin{figure*}
				\center

				\includegraphics[height=4.7cm]{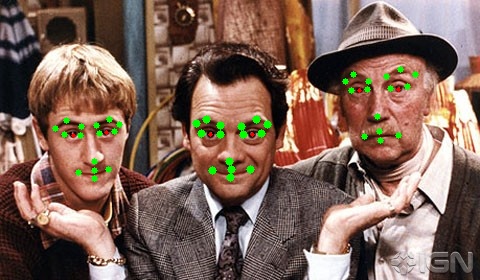}
				\includegraphics[height=4.7cm]{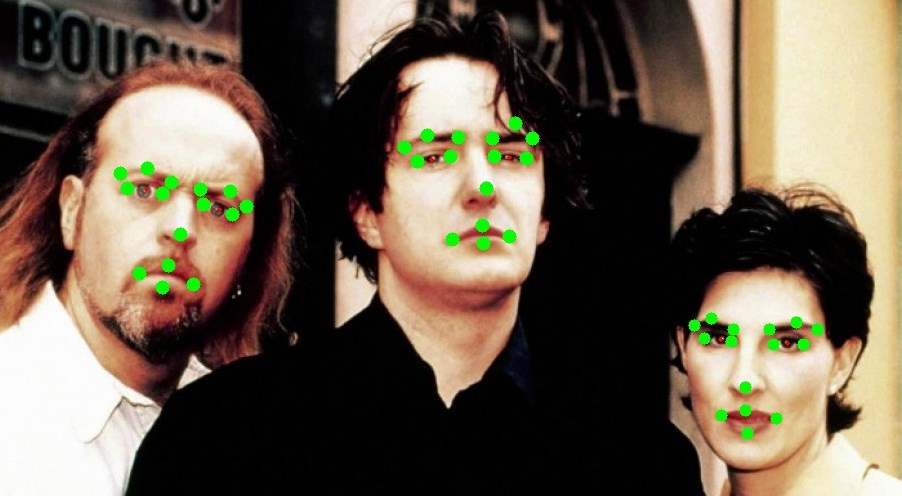}

				\includegraphics[height=5.9cm]{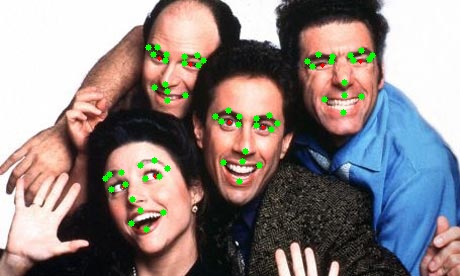}
				\includegraphics[height=5.9cm]{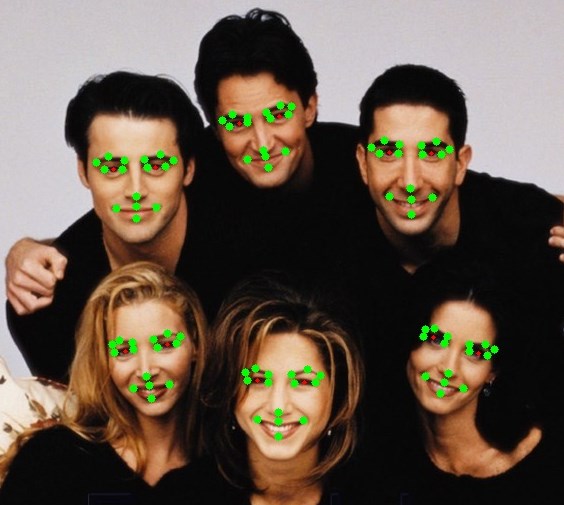}

				\includegraphics[height=5.4cm]{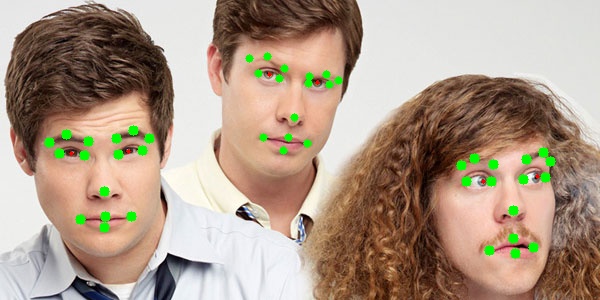}
				\includegraphics[height=5.4cm]{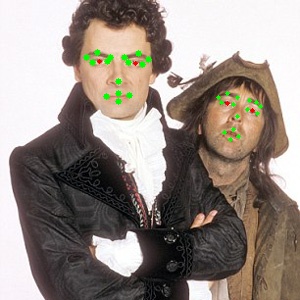}

				\caption
				{
					Some results obtained by our system on real-world images.
					Notice that we use more facial landmark point detectors than in the experimental section.
				}
				\label{fig:qualitative}
			\end{figure*}

			Some qualitative results obtained by our system can be seen in Figure \ref{fig:qualitative} and in the video available at \url{http://youtu.be/xpBXpI39s9c}.
			Furthermore, we prepared a demo application for readers who wish to test the method themselves.
			It is available at \url{http://public.tel.fer.hr/lploc/}.

	\section{Conclusion}
		Numerical results show that our system is less accurate than the reported state-of-the-art but more accurate than two modern commercial products while being considerably faster, in some cases even by a factor of $50$.
		Its landmark point position estimations are, on average, in the pupil diameter ($e\approx 0.05$) from human-annotated ground truth values.
		Processing speed analysis shows that the system can run in real-time on hardware with limited processing power, such as modern mobile devices.
		This enables fast and reasonably accurate facial feature tracking on these devices.
		We believe that the method described in this paper achieves acceptable accuracy and processing speed for a lot of practical applications.

	\section*{Acknowledgements}
		\small
		{
			This research is partially supported by Visage Technologies AB, Link\"{o}ping, Sweden, by the Ministry of Science, Education and Sports of the Republic of Croatia, grant number 036-0362027-2028 "Embodied Conversational Agents for Services in Networked and Mobile Environments" and by the European Union through ACROSS project, 285939 FP7-REGPOT-2011-1.
		}

	\bibliographystyle{IEEEtran}
	\bibliography{references}

\end{document}